\documentclass{ecai}
\usepackage{graphicx}
\usepackage{latexsym}
\usepackage{comment}
\usepackage{algorithmic}
\usepackage{algorithm}
\usepackage{amsfonts}
\usepackage{amsmath}
\usepackage{amssymb}
\usepackage{xcolor}

\usepackage{subcaption}

\begin{document}

\begin{frontmatter}

\title{Counterfactual Prediction Under Selective Confounding}

\author[A,B]{\fnms{Sohaib}~\snm{Kiani}\thanks{Corresponding Author. Email: kianis@beloit.edu}}
\author[C]{\fnms{Jared}~\snm{Barton}}
\author[C]{\fnms{Jon}~\snm{Sushinsky}} 
\author[C]{\fnms{Lynda}~\snm{Heimbach}}
\author[B]{\fnms{Bo}~\snm{Luo}}

\address[A]{Department of Mathematics and Computer Science, Beloit College, USA}
\address[B]{Department of Electrical Engineering and Computer Science, University of Kansas, USA}
\address[C]{School of Social Welfare, University of Kansas, USA}

\begin{abstract}
This research addresses the challenge of conducting interpretable causal inference between a binary treatment and its resulting outcome when not all confounders are known. Confounders are factors that have an influence on both the treatment and the outcome. We relax the requirement of knowing all confounders under desired treatment, which we refer to as \textit{Selective Confounding}, to enable causal inference in diverse real-world scenarios. Our proposed scheme is designed to work in situations where multiple decision-makers with different policies are involved and where there is a re-evaluation mechanism after the initial decision to ensure consistency. These assumptions are more practical to fulfill compared to the availability of all confounders under all treatments.
To tackle the issue of \textit{Selective Confounding}, we propose the use of \textit{dual-treatment} samples. These samples allow us to employ two-step procedures, such as Regression Adjustment or Doubly-Robust, to learn counterfactual predictors. We provide both theoretical error bounds and empirical evidence of the effectiveness of our proposed scheme using synthetic and real-world child placement data. Furthermore, we introduce three evaluation methods specifically tailored to assess the performance in child placement scenarios.
By emphasizing transparency and interpretability, our approach aims to provide decision-makers with a valuable tool. The source code repository of this work is located at \url{https://github.com/sohaib730/CausalML}.

\end{abstract}
\end{frontmatter}

\section{Introduction}
\footnote{Published in European Conference of Artificial Intelligence 2023.}
Accurately predicting the outcomes of treatments is essential for decision-makers in various domains, including health care \cite{health_time}, child welfare \cite{coston_evaluation}, criminal justice \cite{criminal_justice}, and marketing \cite{marketing}. Predictive algorithms that can extract valuable insights from rich case information have the potential to enable better decision-making and improve future outcomes. However, earlier risk assessment tools that rely on past decisions to train predictive algorithms suffer from selective and confounding bias \cite{coston_evaluation}.  In observational data, confounders -- factors that influence both treatment assignment and outcome -- introduce bias in standard supervised learning framework \cite{survey_causal_inference}. To address this issue, counterfactual prediction algorithms have been widely adopted in healthcare \cite{ITE_med} and are now being increasingly used in other domains such as child welfare \cite{Coston_DR} and education \cite{ITE_education,Zheng_decision_making}. For applications involving interventional decisions, counterfactual learning is essential to avoid confounding bias, which can otherwise lead to erroneous training, performance, and fairness evaluations \cite{coston_evaluation}. The problem of counterfactual prediction differs from the standard supervised learning problem \cite{MLvsCausal}. Mainly because outcomes for all treatments are not observed for each unit corresponding to the missing label problem. Unlike traditional prediction models, counterfactual prediction infers causal relationships between treatment and resulting outcomes \cite{pearl_why_causal,MLvsCausal}.

While randomized control trials (RCTs) are viewed as the ideal approach for establishing causal relationships, they are often impractical for numerous applications due to their cost, ethical considerations, or the unavailability of sufficient data \cite{causal_inference_big_data}. Conversely, inferring causality from observed data necessitates the observation of all confounding variables \cite{DR_CB}. This stringent requirement considerably restricts the practicality of these models in real-world scenarios where not all confounding factors are reported or obtainable \cite{Experimental_grounding, Coston_DR}. 

In various decision-making scenarios, such as child placement, ICU admission, or police arrests, limited information is reported or available under the baseline treatment with no intervention. For example, in the context of child welfare, practitioners often face challenges regarding whether to remove children from their homes and place them in foster care or keep them in their homes \cite{child_placement_challenges}. Errors in judgment can occur on either end of this decision spectrum. Children who remain at home may experience additional maltreatment, while children who are removed from their homes may eventually return after a short stay in out-of-home placement before being reunited with their families. These short stays can be traumatic, making it crucial to avoid them \cite{sankaran2018easy}. A counterfactual prediction algorithm can address both issues by answering the question: What would happen if a child is kept in their home? In the field of child welfare, decision-making factors include demographic data, case records, and sensitive information such as maltreatment reports, mental health, and disabilities. It is common for the information reported for children who remain with their families to be less comprehensive compared to those who are placed out of the home \cite{out-of-home-procedure1, out-of-home-procedure2}. This poses a challenge of hidden confounders under the desired treatment, which we refer to as \textit{Selective Confounding}.

Learning causal relationships from observed data is already a complex task, and it becomes even more challenging when confounding factors are hidden \cite{Potential_outcome_framework}. Existing techniques in the literature address this issue in specific settings, such as when hidden confounders are proxies of observed confounders \cite{LatentRep2}, when there is limited randomized control trial (RCT) data available \cite{Experimental_grounding}, or when confounders are hidden only during prediction \cite{Coston_DR}. However, to the best of our knowledge, no method has been proposed that provides an interpretable counterfactual model under the scenario of \textit{Selective Confounding}.

In this work, we address the challenge of \textit{Selective Confounding} using interpretable techniques, relying on two key assumptions: 1) there exists a variation in decision policies during the initial decision-making process. 2) the recidivism of the initial decision in appeal or reevaluation is attributed to errors or biases in the initial judgment, rather than being a consequence of covariate shift. These conditions are applicable in real-world scenarios involving child placement, ICU admission, and police arrests, where multiple decision-makers are involved, and opportunities for appeal or reevaluation exist. For instance, in child placement, we often observe varying decision policies, with initial screenings conducted by police officers or trained child welfare professionals \cite{child_dp}, and state policies varying \cite{varying_dp}. As a result, there are cases of short-stayer children who are reunited with their families soon after being removed due to erroneous out-of-home placements. Similarly, in cases of short stays in the ICU, where patients are released after two days of monitoring, different hospitals or attending physicians correspond to varying decision policies, and the release without any intervention indicates that the stay could have been avoided. Likewise, in instances where individuals are arrested but subsequently released because no charges are filed against them by the attorney, different police officers and locations reflect varying decision policies, and the release without charges may be due to judgmental errors.

This paper proposes a specialized approach that leverages the structure of \textit{selective confounding} by using dual-treatment samples, such as \textit{short stayers} in child placement applications. These samples are unconfounded under \textit{selective confounding} and can be adjusted to obtain the desired treatment outcome, allowing us to identify the target estimator without any confounding bias \cite{Potential_outcome_framework, Coston_DR}. We adapt well-known two-stage learning algorithms, Regression Adjustment (RA) and Doubly Robust (DR) \cite{survey_causal_inference}, to train the target model. In this regard, we have opted for Linear regression as our target model, owing to its intrinsic interpretability, making it highly suitable for sensitive social applications. 

 We demonstrate the effectiveness of our approach through empirical evaluations on both synthetic and real-world data. Since evaluation over real-world data is not straightforward due to missing ground truth labels over subpopulations, we propose novel evaluation methods specifically tailored to child placement applications. Our work represents a significant contribution to the field of causal inference and has the potential to enhance decision-making in sensitive domains, including child welfare and ICU admissions, among others. The complete code for training and reproducing the results over synthetic data is accessible at .

\section{Setup}
\label{sec:setup}
\textbf{Selective Confounding:} Our goal is to predict the outcome under a desired treatment $a_1$ for a binary treatment application where possible treatments are $\{a_1,a_2\}$. Let $X,Z$ be the confounding factors. We have considered the problem setting of \textit{selective confounding} where  $Z$ represents hidden confounders that are not available under desired treatment $T = a_1$ and $X$ are observable under all treatments. Our aim is to relax the condition of observable confounding features to accommodate more applications. According to the potential outcome framework \cite{Potential_outcome_framework}, the target estimator is $\nu_{a_1}(x):= E(Y_{a_1}|X=x)$, where $Y_{a_1} \in R$ is the potential outcome one would observe under treatment $T=a_1$.

\noindent \textbf{Dual-Treatment Samples:} In an observational study of a binary treatment application, few samples experience treatment reversal, e.g., a child reunited with their family after removal or bail may have been granted in the appeal. These samples were previously ignored or considered only under one treatment to ensure training samples are independent. In our work, we still maintain that samples are independent by either including them under one treatment or identifying them with another treatment $T=a_3$ - namely, dual-treatment. We argue that a change in decision usually occurs due to two reasons: 1) Covariate Shift, when things improve or deteriorate over time, or 2) Error in the Initial judgment. The dual-treatment samples are those where decisions have been reverted from $a_2$ to the desired treatment $a_1$ due to a mistake in the initial judgment. For example, in the child placement application, \textit{Short-stayers} are considered dual-treatment samples. Short-stayers are the children who are reunited with their families within 30-days after out-of-home placement \cite{short-stayer}. The 30-day separation window is quite small for a change in covariates, and the main reason behind reunification is errors in earlier judgment \cite{sankaran2018easy}. Thus, the observed treatment for each sample is given as $T \in \{a_1,a_2,a_3\}$.

\noindent \textbf{Data:} The $n$ i.i.d. samples in a training data are described as $D_{i = \{1..n\}} = (X_i,Z_i,T_i,Y_i)$, where $T_i \in \{a_1,a_2,a_3\}$ is the observed treatment, $X_i \in \mathbb{R}^p$ are the observable confounding features, $Z_i \in  \{z : z \in \mathbb{R}^q$ and $T_i \neq a_1 \}$ are the hidden confounding features not reported under treatment $T=a_1$, and $ Y_i \in \mathbb{R} $ is the observed outcome. The impact of hidden confounders is evident through the following propositions:
\begin{list}{}{}
\item \textit{P.2.1 - Training Ignorability:} Decisions are unconfounded given $X$ and $Z$: $Y_{a_2},Y_{a_3} \perp T | X,Z$.

\item \textit{P.2.2 - Selective Confounding:} Decisions are confounded given only $X$: $Y_{a_1} \not\perp T | X$.
\end{list}

\noindent This means that when all confounders are observed, the potential outcomes $Y_{a_2}$ or $ Y_{a_3}$ are directly identifiable from observed data, i.e., $E(Y_{a_2}|X,Z) = E (Y|X,Z,T=a_2)$. Conversely, due to the hidden confounders, the target quantity is not identifiable, i.e., $E(Y_{a_1}|X) \neq E(Y|X,T=a_1)$ \cite{Potential_outcome_framework}.

Next, the marginal and conditional propensity to receive any treatment $t$ are defined as $\pi_{T=t}(X, Z) := P(T=t|X=x, Z=z)$ and $\pi_{T=t|T \neq t'}(X, Z) := P(T=t|X= x, Z=z,T \neq t')$ respectively. The outcome regression for treatment $T=t$ is defined as $\mu_t(X,Z) := E(Y_t|X=x,Z=z)$.

\noindent \textbf{Assumptions:} To identify the target quantity $\nu_{a_1}(x)$, we adopt the following standard assumptions commonly used in causal inference research \cite{pearl_why_causal}, which are easier to satisfy:
\begin{list}{}{}
\item \textit{Consistency:} A case that receives treatment $t$ has outcome $Y = Y_{t}$. This gives $Y= \mathbb{I}(T=a_1)Y_{a_1} + \mathbb{I}(T=a_2)Y_{a_2} + \mathbb{I}(T=a_3)Y_{a_3}$.
\item \textit{Positivity:} There is a non-zero probability of assigning any treatment to a unit, i.e., $\mathbb{P}(\pi_{T=t}(X, Z) \geq \epsilon > 0) = 1$ $\forall\ t \in \{a_1,a_2,a_3\}$.
\end{list}

\noindent \textbf{Miscellaneous notation:} In the paper, we use both $\hat{f}$ and $\tilde{f}$ to denote estimate of $f$, and $\mathbb{I}$ to represent the indicator function. We will use $\mathbb{I}_{t}$ instead of $\mathbb{I}_{T=t}$. For simplicity, we denote the propensity score with short subscripts, such as $\pi_t(.)$ and $\pi_{t|\neq t'}(.)$ for $\pi_{T=t}(.)$ and $\pi_{T=t|T \neq t'}(.)$, respectively. Finally, we represent the target estimator as $\hat{\nu}(x)$ instead of $\hat{\nu}_{a_1}(x)$.

\section{Related Work}
\label{sec:literature}
There has been a significant amount of research on addressing confounding bias with no hidden confounders. One of the classical methods for counterfactual estimation is matching \cite{rubin1973matching,abadie2006large,king2019propensity}. Matching methods estimate the counterfactual outcomes by the nearest neighbor of each individual in terms of covariates. Because the curse of dimensionality makes finding appropriate nearest neighbors of each individual more difficult, propensity score matching, was developed \cite{Propensity_weighting}. Tree-based methods, such as Random forest and Bayesian additive regression trees (BART), have also been applied . \cite{chipman2010bart,hill2011bayesian}. Some literature has focused on learning treatment-invariant representations to remove confounding bias \cite{ITE_RMSE,zhang_latent_ITE}. These methods are not applicable for counterfactual prediction with hidden confounders and will result in more loss \cite{hidden_conf_example1}. 

Methods for addressing the problem of hidden confounders in counterfactual estimation can generally be divided into generic and specialized approaches. Generic approaches tend to focus on robustness and one such approach is sensitivity analysis \cite{Confounding_Policy}. It aims to minimize the worst-case estimated regret of a candidate decision policy compared to a baseline decision policy over an uncertainty set for propensity weights that control the extent of unobserved confounding. However, this approach is unsuitable for our problem settings as it requires observed outcomes from all treatments to estimate the uncertainty set for propensity weights. Another generic approach for addressing hidden confounders is learning latent representations using Variational Autoencoder (VAE) \cite{LatentRep2,Fairness_causal,harada2022infocevae}. These methods assume that observed confounders are proxies of hidden confounders and that hidden confounders can be estimated to some extent through a generative model. The deep generative models such as VAE do not leverage the structure in \textit{selective confounding} to outperform proposed methods. Also, they are not interpretable to be applicable in the social science domain.

In contrast to generic ones, specialized methods exploit the structure in problem in setting to address hidden confounders. Similar to our approach, these methods are more specific to the applications at hand, and provide better performance gains than generic approaches. 

For applications where limited RCT data is available, Kallus et. al. proposed an approach to address hidden confounders \cite{Experimental_grounding}. Since scope of experimental data does not cover observational data, they have proposed to learn calibration parameter from experimental data to adjust counterfactual outcome of observational data. Another approach utilized the network information  to unravel patterns of hidden confounders \cite{ITE_Network_info}.

Structural nested models (SNMs) have been applied in various contexts to address the issue of hidden confounders in data that has multiple levels, such as cross-sectional or longitudinal settings \cite{ SNM_2, SNM_1}. In our case, the focus is on safety when a child stays at home, and not how state-level predictors or programs play any role in the outcome \cite{MLM_guide}. Additionally, the longitudinal aspect of the data is irrelevant to this research question as the aim is not to study how the likelihood of abuse changes over time \cite{MLM_guide}. 

Another widely used approach for addressing hidden confounding in causal inference is the use of Instrument Variables (IV) \cite{IV_1,IV_2,IV_3}. It is important to note that when using IV, certain assumptions must be met such as: 1) the IV being strongly correlated with the treatment variable, 2) independent of the outcome variable and any unobserved confounding factors, and 3) not affecting the outcome variable through any other path than through the treatment variable. In our study, it's not possible that decision policy will always be a valid IV. For instance, in child placement example, a lenient decision policy results in other programs offered by state that ensures In-Home treatment is safe. 

Negative control outcomes are used to address the problem of hidden confounding by providing a way to control for the effects of hidden confounding factors \cite{NCO_1,NCO_2}. The idea is to find an outcome variable that is not affected by the treatment or exposure of interest, but is affected by the same unobserved confounding factors. By comparing the treatment effect on the negative control outcome to the treatment effect on the outcome of interest, it is possible to control for the effects of unobserved confounding factors. These approaches are only suitable for applications where negative control exposures
or outcomes exist. This technique is not suitable for societal applications, as it's not possible to record additional outcomes due to privacy concerns.  

The work closest to ours deals with run-time confounding only \cite{Coston_DR}. For them, training nuisance estimator without confounding bias was straightforward due to no hidden confounders during training. Our work address \textit{selective confounding}, where confounding bias is present during both training and prediction.

\section{Methodology}
\label{sec:algo}
The target model $\hat{\nu}(x)$ is usually termed as a counterfactual predictor \cite{Coston_DR}. Since in an observational study, all samples did not get the desired-treatment to have the factual outcome $Y_{a_1}$.
In this section, we start by introducing the standard method to learn the target model. Then will present the proposed two-step approaches. 
\subsection{Standard Predictor (SP)}
 The straightforward method is to ignore all samples that didn't receive the proposed treatment i.e. with $ T \neq a_1$. This
procedure estimates target quantity $\hat{\nu}(x)$ by learning $\hat{\mathbb{E}}[Y | T = a_1, X = x]$. This method works well-given access to all the confounders i.e. if $T\ \perp Y_{a_1}| X$. However,
under \textit{Selective Confounding} with hidden confounders $Z$, the quantities are not conditional independent. So, the method does not target the right
counterfactual quantity and may produce misleading predictions due to confounding bias \cite{confounding_bias}. 

\subsection{Counterfactual Predictors}
\label{sec:two-step method}
Standard method involves training the model on samples with available ground truth labels $Y_{a_1}$, leading to an estimator $\tilde \omega_{a_1}(x) = \hat{\mathbb{E}}(Y_{a_1}|X)=\hat{\mathbb{E}}(Y|X,T=a_1)$ \cite{standard_practice,coston_evaluation}. However, this approach is only valid under certain conditions, such as when all confounders are observed as given in proposition P.2.1, or in randomized controlled trials (RCT) \cite{Experimental_grounding}. For our problem setting, these conditions are not met, and training on a subpopulation with available ground truth labels will result in underestimating the risk in cases where the desired treatment has historically been assigned to lower-risk cases \cite{Coston_DR}.

To overcome confounding bias during training, we propose leveraging dual-treatment samples, which are defined in Section \ref{sec:setup}. The intuition behind our solution is that dual-treatment samples are unconfounded and it is sometimes feasible to estimate potential outcome $\hat{Y}_{a_1}$ for subpopulation. In that case, the outcome estimator $\hat{\mu}_{a_1}(x,z)=\hat{\mathbb{E}}(\hat{Y}_{a_1}|X,Z) = \hat{\mathbb{E}}(\hat{Y}_{a_1}|X,Z,T=a_3)$ and the conditional propensity score estimator $\hat{\pi}_{a_3|\neq a_1}=\hat{P}(T=a_3|X= x, Z=z,T \neq a_1)$ can be identified without any confounding bias. These two nuisance estimators with no confounding bias can then be used to train the target estimator $\hat{\nu}(x)$ using two-step methods like Regression Adjustment (RA) and Doubly Robust (DR). These two-step methods provide theoretical error bounds and interpretable target model like linear regression \cite{survey_causal_inference}.
\subsubsection{Data Preprocessing - Treatment Adjustment for Dual-Treatment Samples}
\label{sec: data pre-process}
In order to learn target estimator using a two-step procedure, we require potential outcome label $Y_{a_1}$ for dual-treatment samples but the observed label is $Y_{a_3}$ for them. The difference between the outcomes of two treatments is usually given as the Conditional Average Treatment Effect (CATE). By design, samples from desired-treatment $T=a_1$ are closely related to dual-treatment samples. It is possible for certain applications that CATE$(a_1,a_3)$ is negligible. In that case, the observed label $Y_{a_3}$ for dual-treatment samples is equivalent to potential outcome $Y_{a_1}$ and the outcome regressor is identifiable by $\mu_{a_1}(x,z) = \mathbb{E}(Y_{a_1}|X,Z) = \mathbb{E}(Y_{a_3}|X,Z) = \mathbb{E}(Y|X,Z,T=a_3)$.
 \begin{equation*}
     \text{CATE}(a_1,a_3) = \mathbb{E}(Y_{a_1}|X,Z)- \mathbb{E}(Y_{a_3}|X,Z)\\
 \end{equation*}
In case there is a possibility that CATE might be non-zero between two treatments $a_1,a_3$, then it's necessary to estimate potential outcome $\hat{Y}_{a_1}$ for dual-treatment samples. The proposed solution works irrespective of the CATE value.

 \begin{figure}
    \includegraphics[width=\linewidth]{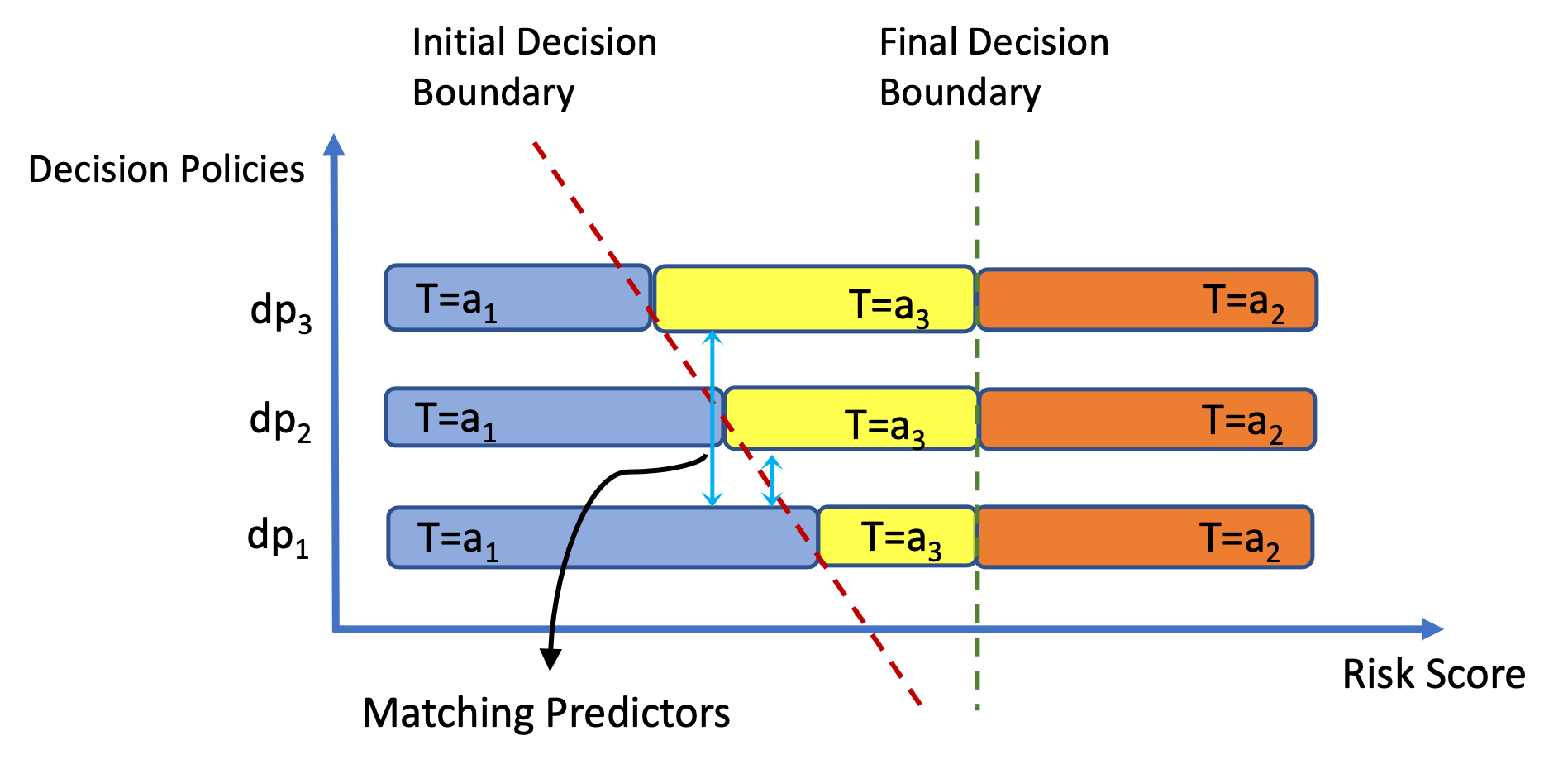}
    \caption{Decision policies across different locations}
    \label{fig:Varying_DP}
 \end{figure}
 
The proposed solution assumes there exist multiple locations with slightly different initial decision policies (dp) as shown in Figure \ref{fig:Varying_DP}.  It represents a real-world scenario when some judges are strict and some are lenient \cite{evaluation_contraction}. The same is the case in child welfare, where some initial placement screenings are performed by police officers or by trained child welfare professionals \cite{child_dp} or state policies are different \cite{varying_dp}.

As mentioned earlier, the regressor $\tilde{\omega}_{a_1}(x) = \hat{\mathbb{E}}(Y |X, T=a_1)$ will be erroneous due to confounding bias. However, in the case of randomized controlled trials (RCT), where $Y_{a_1} \perp T$, there will be no confounding bias for $\tilde{\omega}_{a_1}(x)$ even with hidden confounders \cite{Experimental_grounding}. Although observed samples from the whole population do not result from RCT, in data preprocessing, we only need to estimate $\hat{Y}_{a_1}$ for dual-treatment samples. By definition, dual-treatment samples are those for whom an erroneous decision was made initially and later reverted. In cases where we have varying decision boundaries between locations, the scope of samples with treatment $T=a_1$ from one location covers samples with treatment $T=a_3$ from another location, resembling an RCT between two treatments $a_1$ and $a_3$.

For instance, let's consider there are three different decision policies (DPs) resulting in different decision boundaries for the initial decision as shown in Figure \ref{fig:Varying_DP}.  It shows $dp_1$ is the lenient, and  $dp_3$ is the strictest.  Also, it is expected that the final decision boundary is uniform among locations since the review process is to ensure removing any biases in the judgment process. As a result, the scope of desired-treatment sample $T=a_1$ in $dp_1$ will cover the scope of dual-treatment samples $T=a_3$ in $dp_2$ and $dp_3$. Eventually when samples from all locations are considered as a whole, it will resemble a scenario of RCT between treatments $a_1,a_3$ with $P(T=a_1|X,Z,T \neq a_2)=P(T=a_3|X,Z,T \neq a_2) \approx 0.5$.  

The two subpopulations with treatments $a_1$ and $a_3$ satisfy $Y_{a_1} \perp T | T \neq a_2$. Thus, it is possible to correctly estimate potential outcome  $\hat{Y}_{a_1}$ for dual-treatment samples from $\tilde{\omega}_{a_1}(x)$ . The detailed procedure is provided in Algorithm \ref{alg:calib_pramater_est}. First, it learns $\tilde{\omega}_{a_1}(x) = \hat{\mathbb{E}}(Y|T=a_1,x)$, and then estimates potential outcome $\hat{Y}_{a_1}$ for dual-treatment samples. In Section \ref{sec:synthetic_evaluation}, we have presented the effectiveness of this approach with varying CATE over synthetic data.

\begin{algorithm}
\caption{Label Estimation for dual-treatment samples}
\label{alg:calib_pramater_est}
\begin{algorithmic}
\STATE \textbf{Input:} Data $D_{d \in \{dp_1...dp_L\}} = \{ (X_i,Z_i,T_i,Y_i)\}_{i=1}^{n_d}$
\STATE
\STATE 1. Train regressor $\tilde{\omega}_{a_1}(x)$ over $Y \sim X|T=a_1$ 

\STATE 2. Estimate $\hat{Y}_{a_1}$ from $\tilde{\omega}_{a_1}(x)$ for samples with $T=a_3$ 
\STATE
\STATE \textbf{Return:} Desired Labels $\hat{Y}_{a_1}$ for dual-treatment samples
\end{algorithmic}
\end{algorithm}

\subsubsection{Regression-Adjustment (RA) Predictor}

 In the Regression Adjustment (RA) method, target estimator $\hat{\nu}(x)$ is learned in two stages. The procedure is presented in Algorithm \ref{alg:RA}. First a nuisance estimator $\hat{\mu}_{a_1}(X,Z) = \hat{\mathbb{E}}(\hat{Y}_{a_1}|X,Z,T=a_3)$ is learned with no confounding bias. Then similar to the semi-supervised learning approach the nuisance estimator will be used to provide a pseudo-outcome for the whole population. Then target model $\hat{\nu}(x)$ is estimated over the whole population using ground truth label for samples received treatment $T=a_1$ and pseudo outcome $\hat{\mu}_{a_1}(X,Z)$ for other samples with $T=\{a_2,a_3\}$. 

For theoretical error bounds, we have adopted a procedure similar to \cite{Kennedy2020OptimalDR} which bounds the error for a two-stage regression on the full set of confounding variables.

\begin{theorem}
\label{th:first} Under sample-splitting to learn $\hat{\mu}_{a_1}(x,z)$, $\hat{\nu}(x)$, the RA method has pointwise regression error that is bounded by: 
\begin{multline*}
    \mathbb{E}\Big[\big(\hat{\nu}(x) - \nu(x) \big)^2\Big] \lesssim \mathbb{E} \Big[ \big(\tilde{\nu}(X) - \nu(X)\big)^2\Big] \\
 + \mathbb{E}\Big[\big(\tilde{\mu}_{a_1}(x,Z) - \mu_{a_1}(x,Z)\big)^2|X=x\Big] \\+ \mathbb{E}\Big[\big(\Delta(x,Z)\big)^2|X=x\Big]
\end{multline*}
\end{theorem}

\noindent where $\hat{\nu}(x)$ is the final regressor. The right-hand side splits the error bounds according to the different stages of the algorithm, where $\tilde{\nu}(x) = \hat{\mathbb{E}}(\mathbb{I}_{a_1}Y + \mathbb{I}_{\neq a_1} \mu_{a_1}(X,Z)|X=x)$ denotes second-stage regression with oracle access to the first-stage output. $\tilde{\mu}_{a_1}(X,Z) = \hat{\mathbb{E}}(Y_{a_1}|X,Z)$ is the first stage regressor with oracle access to $Y_{a_1}$ and $\Delta(X,Z) = \mathbb{E}(\hat{Y}_{a_1}-Y_{a_1}|X,Z,T=a_3)$ is the label estimation error for dual-treatment samples. Notably, the bound depend linearly on the Mean square error of $\tilde{\mu}_{a_1}(x,Z)$. The next approach provides a more robust mechanism by introducing another first-stage estimator.

\begin{algorithm}
\caption{Regression-Adjustment (RA) Algorithm}
\label{alg:RA}
\begin{algorithmic}

\STATE \textbf{Input:} Data $ D = \{ (X_i,Z_i,T_i,Y_i)\}_{i=1}^{n}$
\STATE

\STATE {Randomly divide $D$ into two partitions $W^1,W^2$}
\FOR {(i,j) $\in$ \{(1,2),(2,1)\}}
\STATE \textbf{Stage 1:} On $W^i$, learn regression function $\hat{\mu}_{a_1}(X,Z)$ over $\hat{Y}_{a_1} \sim X=x,Z=z|T=a_3$
\STATE
\STATE \textbf{Stage 2:} On $W^j$, learn regressor $\hat{\nu}^j(x)$ over $\mathbb{I}_{a_1}Y + \mathbb{I}_{\neq a_1}\hat{\mu}_{a_1}(X,Z) \sim X$
\ENDFOR
\STATE
\STATE \textbf{Prediction:}  $\hat{\nu}(x) = \sum_{q=1}^2 \hat{\nu}^q(x)$
\end{algorithmic}
\end{algorithm}

\subsubsection{Doubly-Robust (DR) Predictor} This method introduces two nuisance estimators, i.e. $\hat{\mu}_{a_1}(X,Z)$ and $\hat{\pi}_{a_3|\neq a_1}(X,Z)$. It guarantees robustness since it requires only one of the nuisance estimators to be closer to the true estimator. The DR method is presented in Algorithm \ref{alg:DR}. Again it's a two-step procedure. At the first, it trains regressor $\hat{\mu}_{a_1}(X,Z)$ and conditional propensity score model $\hat{\pi}_{a_3|\neq a_1}(X,Z)$. The propensity score model allows us to estimate the likelihood of being a dual-treatment sample given it can not be a desired-treatment sample. Thus to train the propensity score model, no desired-treatment samples with $T = a_1$ are included. The second step of the DR algorithm involves training the target model $\hat{\nu}(x)$ over the whole population. 

DR method provides better theoretical guarantees as compared to RA due to two nuisance estimators.

\begin{theorem}
Under sample-splitting to learn $\hat{\mu}_{a_1}(x,z)$, $\hat{\nu}(x)$ and $\hat{\pi}_{a_3|\neq a_1}(x,z)$ the pointwise error bound for proposed DR method is given as:
\begin{multline*}
\mathbb{E}\Big[\big(\hat{\nu}(x) - \nu(x)\big)^2\Big]
\lesssim \mathbb{E}\Big[\big(\tilde{\nu}(x) - \nu(x)\big)^2\Big]\\
+ \mathbb{E}\Big[\big(\tilde{\mu}_{a_1}(x,Z) - \mu_{a_1}(x,Z)\big)^2|X{=}x\Big] \\
 \times\mathbb{E}\Big[\big
(\hat{\pi}_{a_3|\neq a_1}(x,Z) - \pi_{a_3|\neq a_1}(x,Z)\big)^2|X{=}x\Big] \\+ \mathbb{E}\Big[\big(\Delta(x,Z)\big)^2|X{=}x\Big] 
\end{multline*}
\end{theorem}
\noindent where $\hat{\nu}(x)$ is the final regressor and the error bound is split between different stages at the right-hand side. $\tilde{\nu}(x)$ denotes second-stage regressor with oracle access to true first-stage estimators ${\mu}_{a_1}(x,Z)$, ${\pi}_{a_3|\neq a_1}(x,Z)$. $\tilde{\mu}_{a_1}(x,Z)$ is the first stage regressor with oracle access to $Y_{a_1}$ and $\hat{\pi}_{a_3|\neq a_1}(x,Z)$ is the propensity score estimator. Finally, $\Delta(x,Z) = \mathbb{E}(\hat{Y}_{a_1}-Y_{a_1}|X,Z,T=a_3)$ is the desired label estimation error for dual-treatment samples. Here, in the error bound there is a product between errors of nuisance estimators, which guarantees its robustness. The error bound still linearly depends on the label estimation error of dual-treatment samples. However, that estimation error over subpopulation is much simpler to handle as compared  to the target problem of the whole population. Also, in some cases label estimation for dual-treatment samples might not be needed at all where $\text{CATE}(a_1,a_3)$ is negligible, as mentioned in Section \ref{sec: data pre-process}. 
\begin{algorithm}
\caption{Doubly-Robust (DR) Algorithm}
\label{alg:DR}
\begin{algorithmic}
\STATE \textbf{Input: $ D = \{ (X_i,Z_i,T_i,Y_i)\}_{i=1}^{n}$}
\STATE

\STATE {Randomly divide $D$ into three partitions $W^1,W^2,W^3$}
\FOR {(i,j,k) $\in$ \{(1,2,3), (3,1,2), (2,3,1)\} }
\STATE {\textbf{Stage 1:} On $W^i$, train regression function $\hat{\mu}_{a_1}(X,Z)$ over $\hat{Y}_{a_1} \sim X=x,Z=z|T=a_3$}
\STATE{}
\STATE{\qquad\qquad On $W^j$, train conditional propensity estimator $\hat{\pi}_{a_3|\neq a_1}(X,Z)$ over $\mathbb{I}_{a_3} \sim X=x, Z=z|T \neq a_1$}
\STATE{}
\STATE{\textbf{Stage 2}: On $W^k$, train $\hat{\nu}^k(x)$ by regressing $\mathbb{I}_{a_1}Y + \mathbb{I}_{\neq a_1}\bigg[\frac{\mathbb{I}_{a_3}}{\hat{\pi}_{ a_3|\neq a_1}(X,Z)}\big( \hat{Y}_{a_1} - \hat{\mu}_{a_1}(X,Z)\big)  +  \hat{\mu}_{a_1}(X,Z) \bigg]  \sim X$}
\ENDFOR
\STATE
\STATE \textbf{Prediction:}  $\hat{\nu}(x) = \sum_{q=1}^3 \hat{\nu}^q(x)$

\end{algorithmic}
\end{algorithm}

\section{Experiments}
\label{sec:results}
The evaluation of counterfactual prediction algorithms is challenging due to the absence of ground truth labels for subpopulations with different treatments. To overcome this challenge, counterfactual evaluation is typically performed using synthetic, semi-synthetic, or RCT data \cite{survey_causal_inference, ITE_RMSE}. In this study, we evaluated the proposed counterfactual prediction algorithms using both synthetic and real-world child placement data, and compared the results to the SP baseline approach. It is important to note that we are not aware of any other techniques that are both interpretable and able to predict under \textit{selective confounding}, as discussed in Section \ref{sec:literature}.

\begin{figure*}[htbp]
  \centering
  \begin{minipage}[b]{0.3\textwidth}
    \includegraphics[width=\linewidth]{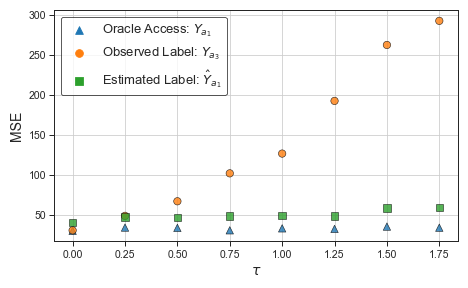}
    \caption{Performance under Varying Treatment Effects}
    \label{fig:figure1}
  \end{minipage}
  \hfill
  \begin{minipage}[b]{0.3\textwidth}
    \includegraphics[width=\linewidth]{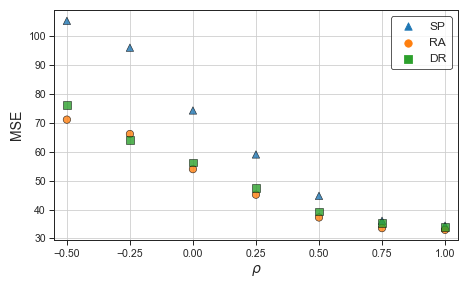}
    \caption{Performance Comparison over Confounders Correlation}
    \label{fig:figure2}
  \end{minipage}
  \hfill
  \begin{minipage}[b]{0.3\textwidth}
    \includegraphics[width=\linewidth]{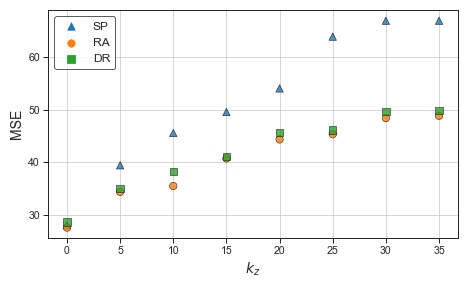}
    \caption{Performance comparison over Hidden-Confounders}
    \label{fig:figure3}
  \end{minipage}
\end{figure*}

\subsection{Synthetic-Data Evaluation}
\label{sec:synthetic_evaluation}

\textbf{Data:} The synthetic data is generated for $L=20$ locations to simulate different decision policies. For each location, the acceptance rate between $ar = [0.3,0.5]$  is randomly picked which will determine how strict or lenient the initial decision policy is. Once determined, the acceptance rate will stay fixed for each location. There is $n=1000$ number of data points in each location.

The confounders $X,Z$ are sampled from Normal Distribution and the correlation among them is controlled using parameter $\rho$. 
\begin{equation*}
    X_i \sim \mathbb{N}(0,1) \ \ , \ Z_i \sim \mathbb{N}(\rho X_i,1-\rho^2)  \ \ \ \ ;0 \leq i \leq d 
\end{equation*}
where $2*d$ is the total number of confounders present in a single sample. The potential outcome $\mu_{a_1},\mu_{a_3}$ under treatments $a_1,a_3$ is determined by:
\begin{align*}
   \mu_{a_1}(X,Z) &= \frac{k_x}{k_x+\rho k_z}\left( \sum_{i=0}^{k_x}X_i + \sum_{i=0}^{k_z}Z_i \right) + \epsilon \\
\mu_{a_3}(X,Z) &=  \frac{k_x}{k_x+\rho k_z}\Big[ \sum_{i=0}^{d}X_i + \sum_{i=0}^{d}Z_i \\
 &- \tau\big ( \sum_{i=0}^{d}X_i + \sum_{i=0}^{d}Z_i^2 \big )\Big] + \epsilon
\end{align*}
with,
\begin{equation*}
    \epsilon \sim \mathbb{N}\big(0,\frac{1}{2n}\left\| \mu_{a_1}(X,Z) \right\|^2_2)
\end{equation*}
and $k_x,k_z$ determines the number of confounders impacting the outcome. CATE between two treatments $a_1,a_3$ is controlled with parameter $\tau$. To compare learned estimator $\hat{\nu}(x)$, the ground truth outcome based on confounders $X$ is given as:
\begin{equation*}
    \nu(X) = \frac{k_x}{k_x+\rho k_z} \Big[ \sum_{i=0}^{k_x}X_i + \rho\sum_{i=0}^{k_z}X_i \Big]
\end{equation*}
The propensity score, probability of treatment assignment, is given as $\pi(X,Z)$. To replicate real-world scenarios where decision reversal is possible through appeals and gives dual-treatment samples, we simulate two decision stages $d_1,d_2$ corresponding to the initial and final decision. 
\begin{align*}
     &\pi(X,Z) = \sigma\Big(\frac{1}{\sqrt{k_x+k_z}}\big(\sum_{i=0}^{k_v}X_i + \sum_{i=0}^{k_z}Z_i \big) \Big )\\
      & d_1 = \text{Bernoulli}(min(\frac{\pi(X,Z)}{\text{ar}}*0.5,0.99))\\
      & d_2 = \text{Bernoulli}(\pi(X,Z))
\end{align*}
where $\sigma(x) = \frac{1}{1+e^{-x}}$ and $ar= [0.3,0.5]$ is the acceptance rate randomly picked for each location in the dataset. Lower $ar$ values mean a strict decision policy as it brings down the decision boundary from $0.5$. Based on $d_1,d_2$, the assigned treatment is given as:
\begin{equation*}
    T= \left\{\begin{matrix}
a_1    & d_1 = 0\\ 
a_2  & d_1 = 1\ \&\ d_2 = 1\\\
a_3 & d_1 = 1\ \&\ d_2 = 0
\end{matrix}\right.
\end{equation*}

\noindent \textbf{Metric:} The dataset  contain samples from all $L$ locations. For evaluation, the dataset is partitioned into train and test sets.  The metric used to evaluate the performance of target estimator $\hat{\nu}(x)$ is Mean Square Error (MSE) given as:
\begin{equation}
    \text{MSE} = \frac{1}{n}\sum_{i = 1}^{n}\big[(\nu(x_i) - \hat{\nu}(x_i))^2\big]
\end{equation}
where $n$ is the total number points of points in the test set.

 \noindent \textbf{Results:} Our solution comprises of two main components: 1) Estimation of potential outcome $\hat{Y}_{a_1}$ for dual-treatment samples using Algorithm \ref{alg:calib_pramater_est} and 2) Training a counterfactual predictor model $\hat{\nu}(x)$ using RA (Algorithm \ref{alg:RA}) and DR (Algorithm \ref{alg:DR}) schemes.

Figure \ref{fig:figure1} demonstrates Algorithm \ref{alg:calib_pramater_est}'s effectiveness by comparing its performance over varying CATE($a_1,a_3$) with three techniques for handling labels in dual-treatment samples. The estimator $\hat{\nu}(x)$ used only the DR scheme to ensure a fair comparison, and setting \textbf{A} in Table \ref{tab:setting} was used for the experiment. The error lower bound is obtained with oracle access to potential outcome $Y_{a_1}$ for dual-treatment samples. Performance deteriorates with increasing $\tau$ when observed label $Y_{a_3}$ for dual-treatment samples is used without label estimation. However, when Algorithm \ref{alg:calib_pramater_est} is used to estimate $\hat{Y}{a_1}$ for dual-treatment samples, performance does not degrade with $\tau$. Moreover, the proposed algorithm \ref{alg:calib_pramater_est} performs similarly to using observed label $Y{a_3}$ as it is when the treatment effect is negligible (i.e., when $\tau \approx 0$). Thus, the proposed algorithm \ref{alg:calib_pramater_est} can be used in any application with different values of $\tau$ without switching between label handling techniques.

\begin{table}
\caption{Experimental settings over synthetic-data} \label{tab:setting}
\begin{center}
\begin{tabular}{llllll}
\textbf{Setting}  &\text{$d$} &\textbf{$k_x$} &\textbf{$k_z$} &\textbf{$\rho$} &\textbf{$\tau$} \\
\hline \\
\textbf{A}         &250 &25 &25 &0.25 &*\\
\textbf{B}             &250 &25 &25 &* &0.5\\
\textbf{C}             &250 &25 &* &0.25 &0.5\\
\end{tabular}
\end{center}
\end{table}

In Figure \ref{fig:figure2}, we compare the final predictor approaches of AR and DR with the baseline SP, using Setting \textbf{B} given in Table \ref{tab:setting}. We plot their performances against the correlation $\rho$ between hidden and observed confounders. As $\rho$ increases, the observed features $X$ serve as proxies for $Z$. In such cases, SP performs comparably due to low confounding bias. However, when the correlation is low, there is a clear advantage of using counterfactual estimation approaches (AR, DR).

Figure \ref{fig:figure3} presents the impact of directly controlling confounding bias using $k_z$, with other parameter values set according to Setting \textbf{C} in Table \ref{tab:setting}. As expected, the performance of all algorithms decreases as the value of $k_z$ increases. However, the proposed counterfactual approaches, AR and DR, outperform the baseline SP, demonstrating their ability to effectively handle confounding bias during training.

It is important to note that in our experiments, there was no clear advantage of using the DR scheme over RA. While the DR scheme involves a more challenging training process for the conditional propensity score estimator ($\hat{\pi}_{{a_3}| \neq {a_1}}(x, z)$) compared to the outcome regressor $\hat{\mu}_{a_1}(x, z)$ used in RA, the DR scheme still provides robustness in case the outcome regressor fails to converge properly.

\subsection{Real-World Data Evaluation}
\textbf{Data:} The real-world data used in this study consists of placement decisions from eight states in the USA, with In-Home placement ($T=a_1$) as the desired treatment. While the eight locations cover a wide area with overlapping attributes, each state may handle cases differently. The observed features $X$ include demographics, case records, and complaints, while hidden features $Z$ include substantiated maltreatment, mental health, and disability type. The relevant outcome is whether or not further maltreatment occurred within six months of In-Home placement.

\begin{figure}[t]
    \includegraphics[width=\linewidth]{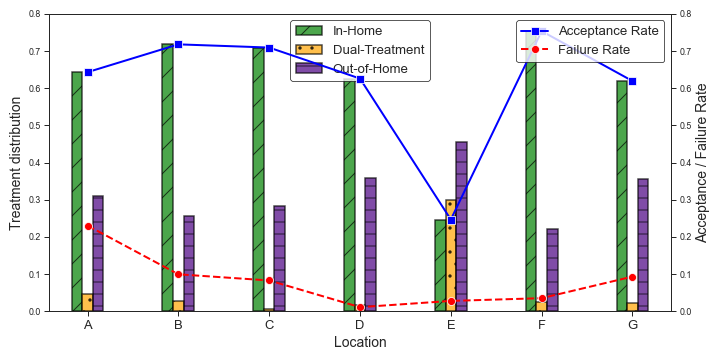}
    \caption{Real-World data Statistics}
    \label{fig:real_dataset}
\end{figure}

Figure \ref{fig:real_dataset} displays $N=5000$ samples randomly drawn from each location. As depicted in Figure \ref{fig:Varying_DP}, nearly similar proportions of Out-of-Home samples at each location led to a uniform final decision boundary. To compare the nature of sub-populations or the effectiveness of initial decision criteria among each state, two new metrics are introduced: Acceptance Rate (AR) and Failure Rate (FR), given as follows:
\begin{align*}
\text{AR} &= \frac{\text{Number of In-Home samples}}{N}\\
\\
\text{FR} &= \frac{\text {Number of maltreated In-Home samples}}{N}
\end{align*}

While a high AR usually results in a high FR and vice versa, not all states follow the same trend, as shown in Figure \ref{fig:real_dataset}. This suggests that In-home sample distribution may vary among states, and certain states may have room for improvement in terms of FR.

\noindent \textbf{Evaluation:} To circumvent the issue of missing ground truth labels in evaluation, we have devised three new methods that are applicable in applications similar to child placement. The first method enables us to quantify the benefit of using a prediction model as compared to historical decision \cite{evaluation_contraction}. Meanwhile, the second and third methods allow us to compare different prediction models to select the best.

\textit{\textbf{Method 1 - Comparison with Humans:}}
In the child placement application, initial decisions may be wrong in two ways 1) higher FR among the In-home population, and 2) significant proportions of dual-treatment samples. For comparison, we see if the proposed prediction model (DR) can switch high-risk in-home samples with low-risk dual-treatment samples. The risk-score for dual-treatment samples can be estimated using Algorithm \ref{alg:calib_pramater_est}.

\begin{figure}[t]
    \includegraphics[width=\linewidth]{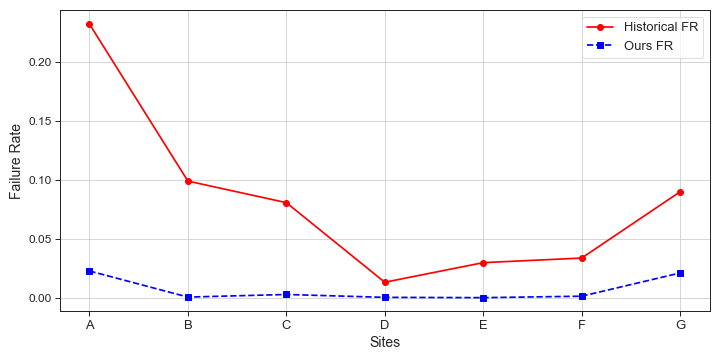}
    \caption{FR Comparison with Similar AR}
    \label{fig:method1}
\end{figure}

 First using DR method, we estimate the risk score for the In-Home and dual-treatment samples. Then without modifying the AR of each location, assign In-Home treatment based on predicted risk scores of In-Home and dual-treatment samples. It can be seen in Figure \ref{fig:method1}, that prediction models bring down the FR for all locations. This improvement in FR is evident either for locations with a high proportion of dual-treatment samples like B,E,F, and G. Or, for a location where FR was already quite poor like A, B, C and G. After further analyses, we have discovered that the most common predictor responsible for \textit{short-stays} is reported as child neglect \cite{sankaran2018easy}, and our prediction algorithm rightfully assigns negative weightage to this predictor.  

\textit{\textbf{Method 2 - Evaluation over Outlier State:}} In Figure \ref{fig:real_dataset}, AR and FR for different states are presented. Two locations, A and E, stand out with minimal difference between AR and FR. For evaluation purposes, location A is important due to its high FR. Its \textit{In-Home} samples contain a significant number of high-risk samples that are usually \textit{Out-of-Home} samples in other states. It allows us to train predictors over other locations B to F and evaluate them over In-Home samples of Location A. The results are reported in Table \ref{tab:RW_Evaluation}, and it can be seen that MSE is different between Standard Predictors (SP) and counterfactual predictors (RA, DR). First, it validates the evaluation method and also shows a gain in performance by circa 10\% due to counterfactual predictors.   

 \begin{table}
\caption{MSE Comparison from Proposed Evaluation Methods} \label{tab:RW_Evaluation}
\begin{center}
\begin{tabular}{lll}
\textbf{Prediction Algorithm}  &\textbf{Method 2} &\textbf{Method 3}\\
\hline \\
\textbf{Standard Predictor (SP)}         &0.8 &0.83  \\
\textbf{Regression Adjustment (RA)}     &0.69 &0.39 \\
\textbf{Doubly-Robust (DR)}             &0.68 &0.26 \\
\end{tabular}
\end{center}
\end{table}

\textit{\textbf{Method 3 - Doubly-Robust Evaluation:}}
The prediction error is identifiable when there are no hidden confounders under desired treatment as $\mathbb{E}((Y_{a_1}-\hat{\nu}(x))^2|X, Z)=\mathbb{E}((\hat{Y}_{a_1}-\hat{\nu}(x))^2|X,Z,T=a_3)$. We propose the Doubly-Robust Evaluation approach similar to \cite{Coston_DR}, with the difference that it can handle \textit{selective confounding}.  Our evaluation procedure identifies error term using dual-treatment samples. Defining identifiable error regressor as $\eta(X,Z):= \mathbb{E}((\hat{Y}_{a_1}-\hat{\nu}(x))^2|X,Z,T=a_3)$, the doubly-robust estimate of MSE is given as:
\begin{multline*}
\frac{1}{n}\sum_{i=1}^{n}\bigg[\mathbb{I}_{\neq a_1}^i\Big[\frac{\mathbb{I}_{a_3}^i}{\hat{\pi}_{ a_3|\neq a_1}(X_i,Z_i)}\big(( \hat{Y}_{a_1}^i - \hat{\nu}(x_i))^2 - \hat{\eta}(X_i,Z_i)\big)\\
+ \hat{\eta}(X_i,Z_i)\Big]+\mathbb{I}_{a_1}^i(Y_i-\hat{\nu}(x_i))^2  \bigg]
\end{multline*}

The results are reported in Table \ref{tab:RW_Evaluation}. It can be seen that there is a significant gain in using counterfactual predictors RA and DR as compared to SP. It is interesting to note that DR performs slightly better than RA which is highlighted by this method.

\section{Summary}
\label{sec:Summary}
Counterfactual predictions for desired treatments play a crucial role in aiding decision-makers in real-world applications. However, assuming all confounders or their proxies are observable might not be realistic. In this work, we address this limitation by considering the scenario of \textit{selective confounding}, where confounders can be hidden under a desired treatment. Despite this relaxation, our proposed method guarantees the identifiability of the target quantity and offers an interpretable target model, subject to two key assumptions: 1) The existence of different decision makers and 2) the possibility of appeal.
To train the target model, we employ a two-stage counterfactual learning technique with no confounding bias.  This allows us to use Linear Regression as the target model, making it inherently interpretable and suitable for societal applications. Our empirical results demonstrate that the proposed framework surpasses both human decision-makers and conventional statistical learning models in terms of performance. While we provide results for child-placement examples, the proposed method can be extended to other applications that satisfy the two conditions, such as ICU admission and police arrests.

\bibliography{ecai-main}
\appendix

%

%

\onecolumn

\section{Appendix}
\subsection{Proof of Theorems}
The theoretical results for our two-stage procedures relies on the theory for pseudo-outcome
regression, which bounds the error for a two-stage regression on the full set of
confounding variables during training \cite{Kennedy2020OptimalDR,Coston_DR}. However, our setting is different due to confounded data in first stage. 
Let $D=(X,Z,T,Y)$ be the training data, split into two parts $D^1$ and $D^2$ to train first stage and second stage estimators respectively. Let $\hat{\mathbb{E}}_n(Y|X=x)$ be the estimator of regression function $\mathbb{E}_n(Y|X=x)$.
The results assume the following two stability conditions
on the second-stage regression estimators:
\begin{itemize}
    \item {Condition 1:} $\hat{\mathbb{E}}_n(Y|X=x) + c \hat{\mathbb{E}}_n(Y+c|X=x)$ for any constant $c$.
    \item {Condition 2:} For any two random variables $R$ and $Q$ if $\mathbb{E}(R|X=x) = \mathbb{E}(Q|X=x)$, then 
\begin{equation*}
    \mathbb{E}\Big[\big(\hat{\mathbb{E}}_n(R|X=x) - {\mathbb{E}}(R|X=x)\big)^2\Big] \asymp \mathbb{E}\Big[\big(\hat{\mathbb{E}}_n(Q|X=x) - {\mathbb{E}}(Q|X=x)\big)^2\Big]
\end{equation*}
\end{itemize}

\noindent In our notations ${m}(x)= \mathbb{E}(f(Z)|X=x)$ represents the conditional expectation. $\tilde{m}(x)= \hat{\mathbb{E}}(f(Z)|X=x)$ represents oracle regression. $\hat{m}(x)= \hat{\mathbb{E}}(\hat{f}(Z)|X=x)$ represents regression over estimated $\hat{f}(Z)$.

\noindent Finally, the two-stage regressor error bound can be written as \cite{Kennedy2020OptimalDR}:
\begin{equation*}
    \mathbb{E}\Big[\big(\hat{\nu}(x) - \nu(x) \big)^2\Big] \lesssim \mathbb{E} \Big[ \big(\tilde{\nu}(X) - \nu(X)\big)^2\Big] \\
 + \mathbb{E}\Big[\hat{r}(x)^2\Big] 
\end{equation*}

\textbf{Theorem 1:}
\begin{align*}
    \hat{r}(x)&=\mathbb{E}_{D|X=x}\big( \mathbb{I}_{a_1}.Y + \mathbb{I}_{\neq a_1}.\hat{\mu}_{a_1}(x,Z) \big) - \mathbb{E}\big( \mu_{a_1}(x,Z)|X=x\big)\\
&= \mathbb{E}_{Z,T|X=x}\Big[E_{Y|T=t,X=x,Z=z} \big(\mathbb{I}_{a_1}.Y + \mathbb{I}_{\neq a_1}.\hat{\mu}_{a_1}(x,Z)\big) \Big] - \mathbb{E}\big( \mu_{a_1}(x,Z)|X=x\big)\\
&= \mathbb{E}_{Z,T|X=x}\Big[ \big(\mathbb{I}_{a_1}.\mu_{a_1}(x,Z) + \mathbb{I}_{\neq a_1}.\hat{\mu}_{a_1}(x,Z)\big) \Big] - \mathbb{E}\big( \mu_{a_1}(x,Z)|X=x\big)\\
    &=\mathbb{E}_{Z,T|X=x}\Big[ \big(\mathbb{I}_{a_1}.\mu_{a_1}(x,Z) + \mathbb{I}_{\neq a_1}.\hat{\mu}_{a_1}(x,Z) \big) \Big] - \mathbb{E}\big( \mathbb{I}_{a_1}\mu_{a_1}(x,Z) + \mathbb{I}_{\neq a_1} \mu_{a_1}(x,Z)|X=x\big)\\
&=\mathbb{E}_{Z,T|X=x}\Big[ \big(\mathbb{I}_{a_1}.\mu_{a_1}(x,Z) + \mathbb{I}_{\neq a_1}.\hat{\mu}_{a_1}(x,Z) - \mathbb{I}_{a_1}\mu_{a_1}(x,Z) - \mathbb{I}_{\neq a_1} \mu_{a_1}(x,Z)\big) \Big]\\
&=\mathbb{E}_{Z,T|X=x}\Big[ \big( \mathbb{I}_{\neq a_1}.\hat{\mu}_{a_1}(x,Z) - \mathbb{I}_{\neq a_1} \mu_{a_1}(x,Z)\big) \Big]\\
     &=\mathbb{E}_{Z,T|X=x}\Big[ \big( \mathbb{I}_{\neq a_1}.\hat{\mathbb{E}}_{n}(\hat{Y}_{a_1}|X=x,Z=z) - \mathbb{I}_{\neq a_1} \mu_{a_1}(x,Z)\big) \Big]\\
&=\mathbb{E}_{Z,T|X=x}\Big[ \big( \mathbb{I}_{\neq a_1}.\hat{\mathbb{E}}_{n}(Y_{a_1} + \hat{Y}_{a_1} - Y_{a_1}|X=x,Z=z) - \mathbb{I}_{\neq a_1} \mu_{a_1}(x,Z)\big) \Big]\\
&=\mathbb{E}_{Z,T|X=x}\Big[ \mathbb{I}_{\neq a_1}\big(\tilde{\mu}_{a_1}(x,Z) +\Delta(x,Z) - \mu_{a_1}(x,Z)\big) \Big]\\
&=\mathbb{E}_{Z|X=x}\Big[ \pi_{\neq a_1}(x,Z)\big(\tilde{\mu}_{a_1}(x,Z) +\Delta(x,Z) - \mu_{a_1}(x,Z)\big) \Big]\\
&\leq \mathbb{E}\Big[\big(\tilde{\mu}_{a_1}(x,Z) - \mu_{a_1}(x,Z)\big)|X=x \Big] + \mathbb{E}\Big[\Delta(x,Z)|X=x\Big]\\
 \end{align*}
The first line is the definition of error term $\hat{r}$ and the second line by iterated expectation. The third line uses the fact that only randomness inside is $Y$ and uses the definition of $\mu_{a_1}(x,Z) = \mathbb{E}(Y|X,Z,T=a_1)$. The fourth line split second term among two populations. The fifth and sixth line rearranges terms and performs cancellation. The seventh line uses definition of $\hat{\mu}_{a_1}(x,z)$. Eighth line add and subtract $Y_a1$. The ninth line use definitions of $\tilde{\mu}_{a_1}(x,Z)$ and   $\Delta(x,Z) := \hat{Y}_{a_1} - Y_{a_1}$. The tenth line performs iterated expectation $E_{T|X=x,Z=z}$ inside, where $T$ is the only random variable and applies definition of $\pi_{\neq a_1}(x,Z) = P(T \neq a_1|X,Z)$. The eleventh line ignores $\pi_{\neq a_1}(x,Z)$ to get the upper bound and rearranges the terms. 

\noindent Taking square on both sides and using the fact $(a+b)^2 \leq a^2 + b^2$ gives:
\begin{equation*}
   \hat{r}(x)^2 \leq \mathbb{E}\Big[\big(\tilde{\mu}_{a_1}(x,Z) - \mu_{a_1}(x,Z)\big)^2|X=x \Big] + \mathbb{E}\Big[\Delta(x,Z)^2|X=x\Big]
\end{equation*}
If $\hat{\mu}$ are estimated using separate training samples $D^1$, then taking the expectation over
the first-stage training sample yields:
\begin{equation*}
    \mathbb{E}(\hat{r}(x)^2)=\mathbb{E}\Big[\big(\tilde{\mu}_{a_1}(x,Z) - \mu_{a_1}(x,Z)\big)^2|X=x \Big] + \mathbb{E}\Big[\Delta(x,Z)^2|X=x\Big]
\end{equation*}

\textbf{Theorem 2: }
\begin{align*}
    \hat{r}(x)&=\mathbb{E}_{D|X=x}\Big[ \mathbb{I}_{a_1}.Y + \mathbb{I}_{\neq a_1}.\big\{\frac{\mathbb{I}_{a_3}}{\hat{\pi}_{a_3|\neq a_1}(x,Z)}(\hat{Y}_{a_1}-\hat{\mu}_{a_1}(x,Z)) + \hat{\mu}_{a_1}(x,Z) \big\}\Big] \\
    &\qquad \qquad \qquad \qquad \qquad \qquad \qquad \qquad \qquad \qquad \qquad \qquad - \mathbb{E}\big( \mathbb{I}_{a_1}\mu_{a_1}(x,Z) + \mathbb{I}_{\neq a_1}\mu_{a_1}(x,Z)|X=x\big)\\
&= \mathbb{E}_{Z,T|X=x}\Big[E_{Y|T=t,X=x,Z=z}\ \ \mathbb{I}_{\neq a_1}.\big\{\frac{\mathbb{I}_{a_3}}{\hat{\pi}_{a_3|\neq a_1}(x,Z)}(Y_{a_1} + \Delta(x,Z) - \hat{\mu}_{a_1}(x,Z)) + \hat{\mu}_{a_1}(x,Z) - \mu_{a_1}(x,Z)\big\}\Big]\\
&= \mathbb{E}_{Z,T|X=x}\Big[\mathbb{I}_{\neq a_1}.\big\{\frac{\mathbb{I}_{a_3}}{\hat{\pi}_{a_3|\neq a_1}(x,Z)}(\mu_{a_1}(x,Z) - \hat{\mu}_{a_1}(x,Z) + \Delta(x,Z)) + \hat{\mu}_{a_1}(x,Z) - \mu_{a_1}(x,Z) \big\}\Big]\\
&= \mathbb{E}_{Z|X=x}\Big[\pi_{\neq a_1}(x,Z).\big\{\frac{\pi_{a_3|\neq a_1}(x,Z)}{\hat{\pi}_{a_3|\neq a_1}(x,Z)}(\mu_{a_1}(x,Z) - \hat{\mu}_{a_1}(x,Z)+ \Delta(x,Z)) + \hat{\mu}_{a_1}(x,Z) - \mu_{a_1}(x,Z) \big\}\Big]\\
    &= \mathbb{E}_{Z|X=x}\Big[\pi_{\neq a_1}(x,Z).\big\{\frac{1}{\hat{\pi}_{a_3|\neq a_1}(x,Z)}(\mu_{a_1}(x,Z) - \hat{\mu}_{a_1}(x,Z))(\pi_{a_3|\neq a_1}(x,Z) - \hat{\pi}_{a_3|\neq a_1}(x,Z))\\
    &\qquad + \frac{\pi_{a_3|\neq a_1}(x,Z)}{\hat{\pi}_{a_3|\neq a_1}(x,Z)}\Delta(x,Z)\big\}\Big]\\
&\leq \mathbb{E}_{Z|X=x}\Big[\big\{\frac{1}{\hat{\pi}_{a_3|\neq a_1}(x,Z)}(\mu_{a_1}(x,Z) - \hat{\mu}_{a_1}(x,Z))(\pi_{a_3|\neq a_1}(x,Z) - \hat{\pi}_{a_3|\neq a_1}(x,Z))\big\}\Big] \\
&\qquad \qquad \qquad \qquad \qquad \qquad \qquad \qquad \qquad \qquad \qquad \qquad + \mathbb{E}_{Z|X=x}\Big[\frac{\pi_{a_3|\neq a_1}(x,Z)}{\hat{\pi}_{a_3|\neq a_1}(x,Z)}\Delta(x,Z)\Big]\\
\end{align*}
The first line uses the definition of $\hat{r}$. The second line rearrange terms, replace $\hat{Y}_{a_1}$ with $\hat{Y}_{a_1}+Y_{a_1}-Y_{a_1}$ and apply iterated expectation. In the third line only randomness inside is $Y$ and uses the definition of $\mu_{a_1}$. Fourth line uses iterated expecatation over $T$ and applies definition of $\pi_t$. The fifth and sixth lines simply rearrange terms. 

\noindent By Cauchy-Schwarz Inequality and positivity assumption, we get:
\begin{align*}
    \hat{r}(x) \leq  C.\sqrt{\mathbb{E}[(\mu_{a_1}(x,Z) - \hat{\mu}_{a_1}(x,Z))^2|X=x]}\sqrt{\mathbb{E}[(\pi_{a_3|\neq a_1}(x,Z) - \hat{\pi}_{a_3|\neq a_1}(x,Z))^2|X=x]} + \mathbb{E}[\Delta(x,Z)|X=x]
\end{align*}
where $C$ is constant. Then take square on both sides, and use the fact $(a+b)^2 \leq a^2 + b^2$ and apply Jensen's Inequality on the second term.
\begin{equation*}
  \hat{r}(x)^2 \leq  {\mathbb{E}[(\mu_{a_1}(x,Z) - \hat{\mu}_{a_1}(x,Z))^2|X=x]}.{\mathbb{E}[(\pi_{a_3|\neq a_1}(x,Z) - \hat{\pi}_{a_3|\neq a_1}(x,Z))^2|X=x]} + \mathbb{E}[\Delta(x,Z)^2|X=x]\\  
\end{equation*}

\noindent If nuisance estimator $\hat{\pi}$ and $\hat{\mu}$ are estimated using separate training samples, then taking the expectation over
the first-stage training sample yields:
\begin{equation*}
    \mathbb{E}(\hat{r}(x)^2) \leq  {\mathbb{E}[(\mu_{a_1}(x,Z) - \hat{\mu}_{a_1}(x,Z))^2|X=x]}.{\mathbb{E}[(\pi_{a_3|\neq a_1}(x,Z) - \hat{\pi}_{a_3|\neq a_1}(x,Z))^2|X=x]} + \mathbb{E}[\Delta(x,Z)^2|X=x]\\
\end{equation*}

\end{document}